444# Probabilistic evaluation of sequential plans from causal models with hidden variables

**Judea Pearl**
Cognitive Systems Laboratory
Computer Science Department
University of California, Los Angeles
Los Angeles, CA 90095-1596
*judea@cs.ucla.edu*

**James Robins**
Departments of Epidemiology & Biostatistics
School of Public Health
Harvard University
Boston, MA 02115
*robins@hsph.harvard.edu*

## Abstract

The paper concerns the probabilistic evaluation of plans in the presence of unmeasured variables, each plan consisting of several concurrent or sequential actions. We establish a graphical criterion for recognizing when the effects of a given plan can be predicted from passive observations on measured variables only. When the criterion is satisfied, a closed-form expression is provided for the probability that the plan will achieve a specified goal.

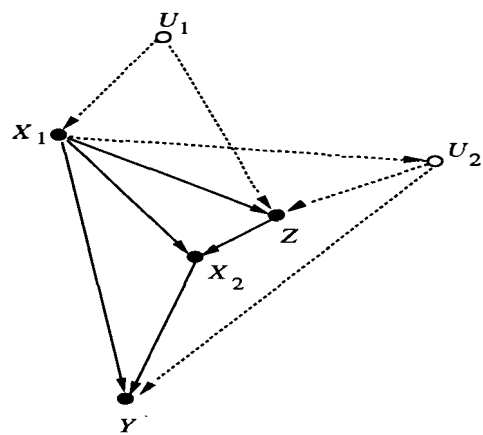

*Figure 1*: Illustrating the problem of evaluating the effect of the plan $(do(x_1), do(x2))$ on $Y$, from nonexperimental data taken on $X_1$, $Z$, $X_2$ and $Y$.

## 1 INTRODUCTION

The problem addressed in this paper is the probabilistic evaluation of the effects of plans, when knowledge is encoded in the form of a partially specified causal diagram. We are given the topology of the diagram but not the conditional probabilities on all variables. Numerical probabilities are given to only a subset of variables which are deemed "observable", while those deemed "unobservable" serve only to specify possible connections among observed quantities, but are not given numerical probabilities.

To motivate the discussion, consider an example discussed in Robins (1993, Appendix 2), as depicted in Figure 1. The variables $X_1$ and $X_2$ stand for treatments that physicians prescribe to a patient at two different times, $Z$ represents observations that the second physician consults to determine $X_2$, and $Y$ represents the patient's survival. The hidden variables $U_1$ and $U_2$ represent, respectively, part of the patient history and the patient disposition to recover. A simple realization of such structure could be found among AIDS patients, where $Z$ represents episodes of PCP – a common opportunistic infection of AIDS patients which, as the diagram shows, does not have a direct effect on survival ($Y$) (since it can be treated effectively) but is an indicator of the patient's underlying immune status ($U_2$) which can cause death ($Y$). $X_1$ and $X_2$ stand for bactrim – a drug that prevents PCP ($Z$) and may also prevent death by other mechanisms. Doctors used the patient's earlier PCP history ($U_1$) to prescribe $X_1$, but its value was not recorded for data analysis.

The problem we face is as follows. Assume we have collected a large amount of data on the behavior of many patients and physicians, which is summarized in the form of (an estimated) joint distribution $P$ of the observed four variables ($X_1$, $Z$, $X_2$, $Y$). A new patient comes in and we wish to determine the impact of the (unconditional) plan $(do(x_1), do(x_2))$ on survival ($Y$), where $x_1$ and $x_2$ are two predetermined dosages of bactrim, to be administered at two prespecified times.

More generally, our problem amounts to that of evaluating a new plan by watching the performance of other planners whose decision strategies are indiscernible. Physicians do not provide a description of all inputs which prompted them to prescribe a given treatment; all they communicate to us is that $U_1$ was consulted in determining $X_1$ and that $Z$ and $X_1$ were consulted in determining $X_2$. But $U_1$, unfortunately, was not recorded.

The problem of learning from the performance of other planners is that one is never sure whether an observed response is due to the planner's action or due to the event which triggered that action and simultaneously caused the response. Such events are called "confounders". The standard techniques of dealing with potential confounders is to adjust for possible variations of confounders by stratification. This amounts to conditioning the distribution on the various states of the confounding variables, evaluating the effect of the plan in each state separately, then taking the (weighted ) average over those states. However, in planning problems like the one above stratification is exacerbated by two problems. First, some of the potential confounders are unobservable (e.g., $U_1$), so they cannot be conditioned on. Second, some of the confounders (e.g., $Z$) are affected by the control variables and, one of the deadliest sins in the design of statistical experiments [Cox 1958, page 48] is to stratify on such variables. The sin being that stratification simulates holding a variable constant, but holding constant a variable that stands between an action and its consequence prevents us from obtaining an accurate reading on the unmediated effect of that action.

The techniques developed in this paper will enable us to recognize in general, by graphical means, whether a proposed plan can be evaluated from the joint distribution on the observables and, if the answer is positive, which covariates should be adjusted for, and how.

Our starting point is a knowledge specification scheme in the form of a causal diagram, like the one shown in Figure 1, which provides a qualitative summary of the analyst's understanding of the relevant data-generating processes.[1] The semantics behind causal diagrams and their relations to actions and belief networks have been discussed in prior publications [Pearl & Verma 1991, Goldszmidt & Pearl 1992, Druzdzel & Simon 1993, Pearl 1993a, Spirtes et al. 1993, Pearl 1993b]. In Spirtes et al. (1993) and later in Pearl (1993b), for example, it was shown how causal networks can be used to facilitate quantitative predictions of the effects of interventions, including interventions that were not contemplated during the network's construction. A more recent paper [Pearl 1994] reviews this aspect of causal networks, and proposes a calculus for deriving probabilistic assessments of the effects of actions in the presence of unmeasured variables. Using this calculus (reviewed in Appendix I) graphical criteria can be established for deciding whether the effect of one variable ($X$) on another ($Y$) is *identifiable* from sample data involving only observed variables, namely, whether it is possible to extract from such data a consistent estimate of the probability of $Y$ under hypothetical interventions with variables $X$. In a related paper [Galles & Pearl 1995] it is shown that the identification of causal effect between two singleton variables (say $X_1$ and $Y_1$) can be accomplished systematically, in time polynomial in the number of variables in the graph.

This paper extends certain results of [Galles & Pearl 1995] to the case where $X$ stands for a compound action, consisting of several atomic interventions which are implemented either concurrently or sequentially. We establish a graphical criterion for recognizing when the effect of $X$ on $Y$ is identifiable and, in case the diagram satisfies this criterion, we provide a closed-form expression for the distribution of an outcome variable $Y$ under the plan defined by the compound action $do(X = x)$. The derived expressions invoke only measured probabilities as obtained, for example, by recording past performances of other planning agents or, in case the elements of $X$ are not controlled by agents, by taking passive measurements from the environment. If $Y$ stands for a goal variable, then the formula provides an expression for the probability that the plan $X$ would lead to goal satisfaction.

## 2 PLAN IDENTIFICATION

**Notation:**
A *control problem* consists of a directed acyclic graph (DAG) $G$ with vertex set $V$, partitioned into four disjoint sets $V = \{X, Z, U, Y\}$

- $X$ – represents the set of control variables (exposures, interventions, etc.)

- $Z$ – represents the set of observed variables, often called *covariates*.

- $U$ – represents the set of unobserved (latent) variables.

- $Y$ – represents an outcome variable.

In this section, we let the control variables be temporally ordered $X = X_1, X_2, \ldots, X_n$ such that every $X_k$ is an ancestor of $X_{k+j}(j > 0)$ in $G$, and we let the outcome $Y$ be a descendent of $X_n$. We relax these assumptions in Section 6. Let $N_k$ stand for the set of observed nodes that are nondescendents of $X_k$. A **plan** is an ordered sequence $(\hat{x}_1, \hat{x}_2, \ldots, \hat{x}_n)$ of value-assignments to the control variables, where $\hat{x}_k$ means "$X_k$ is set to $x_k$". A **conditional plan** is an ordered sequence $(\hat{g}_1(z), \hat{g}_2(z), \ldots, \hat{g}_n(z))$ where each $g_k$ is a function from $Z$ to $X_k$, and $\hat{g}_k(z)$ stands for the statement "set $X_k$ to $g_k(z)$ whenever $Z$ attains the value $z$". The support of each $g_k(z)$ function must not contain any $Z$ variables which are descendents of $X_k$ in $G$.

---
[1] An alternative specification scheme using counterfactual statements was developed earlier by Robins (1986, 1987), and was used to study the identification problem by non-graphical techniques. Robins' scheme extended Rubin's (1978) counterfactual scheme for singleton actions to compound actions and plans.

Our problem is to **evaluate** an unconditional plan[2], namely, to compute $P(y|\hat{x}_1, \hat{x}_2, \ldots, \hat{x}_n)$ which represents the impact of the plan $(\hat{x}_1, \ldots, \hat{x}_n)$ on the outcome variable $Y$. The expression $P(y|\hat{x}_1, \hat{x}_2, \ldots, \hat{x}_n)$ is said to be **identifiable** in $G$ if, for every assignment $(\hat{x}_1, \hat{x}_2, \ldots, \hat{x}_n)$, the expression can be determined uniquely from the joint distribution of the observables $\{X, Y, Z\}$. A control problem is said to be identifiable whenever $P(y|\hat{x}_1, \hat{x}_2, \ldots, \hat{x}_n)$ is identifiable.

Our main identifiability criteria are presented in Theorems 1 and 3 below. These invoke $d$-separation tests on various subgraphs of $G$, defined as follows. Let $X, Y$, and $Z$ be arbitrary disjoint sets of nodes in a DAG $G$. We use the expression $(X \perp\!\!\!\perp Y|Z)_G$ to denote that the set $Z$ $d$-separates $X$ from $Y$ in $G$. We denote by $G_{\overline{X}}$ the graph obtained by deleting from $G$ all arrows pointing to nodes in $X$. Likewise, we denote by $G_{\underline{X}}$ the graph obtained by deleting from $G$ all arrows emerging from nodes in $X$. To represent the deletion of both incoming and outgoing arrows, we use the notation $G_{\overline{X}\underline{Z}}$. Finally, the expression $P(y|\hat{x}, z) \triangleq P(y, z|\hat{x})/P(z|\hat{x})$ stands for the probability of $Y = y$ given that $Z = z$ is observed and $X$ is held constant at $x$.

## 3  ADMISSIBLE MEASUREMENTS

**Theorem 1** $P(y|\hat{x}_1, \ldots, \hat{x}_n)$ *is identifiable if for every* $1 \leq k \leq n$ *there exists a set* $Z_k$ *of covariates satisfying:*

$$Z_k \subseteq N_k, \qquad (1)$$

*(i.e., $Z_k$ consists of non-descendants of $X_k$) and*

$$(Y \perp\!\!\!\perp X_k | X_1, \ldots, X_{k-1}, Z_1, Z_2, \ldots, Z_k)_{G_{\underline{X_k}, \overline{X}_{k+1}, \ldots, \overline{X}_n}} \qquad (2)$$

*When the conditions above are satisfied, the plan evaluates to*[3]

$$P(y|\hat{x}_1, \ldots, \hat{x}_n) = \sum_{z_1, \ldots, z_n} P(y|z_1, \ldots, z_n, x_1, \ldots, x_n) \prod_{k=1}^{n} P(z_k|z_1, \ldots, z_{k-1}, x_1, \ldots, x_{k-1}) \qquad (3)$$

Before presenting its proof, let us demonstrate how Theorem 1 can be used to test the identifiability of the control problem shown in Figure 1. First, we will show that $P(y|\hat{x}_1, \hat{x}_2)$ cannot be identified without measuring $Z$, namely, the sequence $Z_1 = \{\emptyset\}, Z_2 = \{0\}$ would not satisfy conditions (1)-(2). The two $d$-separation tests encoded in (2) are:

$$(Y \perp\!\!\!\perp X_1)_{G_{\underline{X}_1, \overline{X}_2}} \text{ and } (Y \perp\!\!\!\perp X_2|X_1)_{G_{\underline{X}_2}}$$

The two subgraphs associated with these tests are shown in Figure 2, (a) and (b), respectively. We see that, while $(Y \perp\!\!\!\perp X_1)$ holds in $G_{\underline{X}_1, \overline{X}_2}$, $(Y \perp\!\!\!\perp X_2|X_1)$ fails to hold in $G_{\underline{X}_2}$. Thus, in order to pass the test, we must have either $Z_1 = \{Z\}$ or $Z_2 = \{Z\}$ but, since $Z$ is a descendant of $X_1$, only the second alternative remains: $Z_2 = \{Z\}$. The test applicable to the sequence $Z_1 = \{\emptyset\}, Z_2 = \{Z\}$ are: $(Y \perp\!\!\!\perp X_1)_{G_{\underline{X}_1, \overline{X}_2}}$ and $(Y \perp\!\!\!\perp X_2|X_1, Z)_{G_{\underline{X}_2}}$. Figure 2 shows that both tests are now satisfied, because $\{X_1, Z\}$ $d$-separates $Y$ from $X_2$ in Figure 2(b). Having satisfied conditions (1)-(2), Eq. (3) provides a formula for the effect of plan $(\hat{x}_1, \hat{x}_2)$ on $Y$:

$$P(y|\hat{x}_1, \hat{x}_2) = \sum_z P(y|z, x_1, x_2) P(z|x_1) \qquad (4)$$

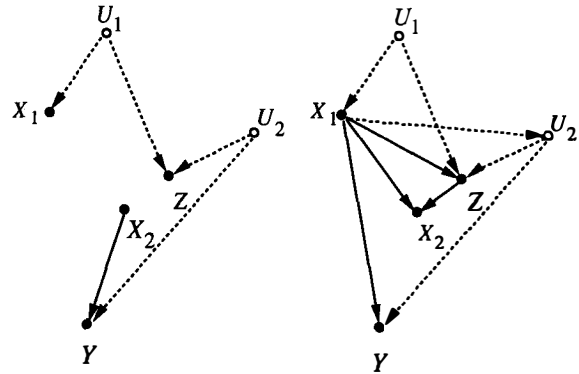

*Figure 2*: The two subgraphs of $G$ used in testing the identifiability of $(\hat{x}_1, \hat{x}_2)$; (a) $G_{\underline{X}_1, \overline{X}_2}$ and (b) $G_{\underline{X}_2}$.

The question naturally arises whether the sequence $Z_1 = \{\emptyset\}, Z_2 = \{Z\}$ can be identified without exhaustive search. This will be answered in Corollary 2 and Theorem 3.

**Proof of Theorem 1:** The proof given here is based on the inference rules described in Appendix I which facilitate the reduction of causal-effect formulas to hat-free expressions. An alternative proof is provided in Section 6.1.

1. The condition $Z_k \subseteq N_k$ implies $Z_k \subseteq N_j$ for all $j \geq k$. Therefore, we have

$$P(z_k|z_1, \ldots, z_{k-1}, x_1, \ldots, x_{k-1}, \hat{x}_k, \hat{x}_{k+1}, \ldots, \hat{x}_n) = P(z_k|z_1, \ldots, z_{k-1}, x_1, \ldots, x_{k-1})$$

---
[2]Identification of conditional plans has been considered in [Robins 1986, 1987] and certain extensions of our graphical results are presented in [Pearl 1994, Robins & Pearl 1995].

[3]The computation and estimation of sum-product expressions of the form given in Eq. (3), where $Z_k$ stand for any subset of $N_k$, were investigated by J. Robins under the rubric "G-computation algorithm formula" [Robins 1986], hereafter G-formula.

This is so because no node in $\{Z_1, \ldots, Z_k, X_1, \ldots, X_{k-1}\}$ can be a descendant of any node in $\{X_k, \ldots, X_n\}$, hence, Rule 3 allows us to delete the hat variables from the expression.

2. Condition (2) permits us to invoke Rule 2 and write:

$$P(y|z_1, \ldots, z_k, x_1, \ldots, x_{k-1}, \hat{x}_k, \hat{x}_{k+1}, \ldots, \hat{x}_n) =$$
$$P(y|z_1, \ldots, z_k, x_1, \ldots, x_{k-1}, x_k, \hat{x}_{k+1}, \ldots, \hat{x}_n)$$

Thus, we have

$$P(y|\hat{x}_1, \ldots, \hat{x}_n)$$
$$= \sum_{z_1} P(y|z_1, \hat{x}_1, \hat{x}_2, \ldots, \hat{x}_n) P(z_1|\hat{x}_1, \ldots, \hat{x}_n)$$
$$= \sum_{z_1} P(y|z_1, x_1, \hat{x}_2, \ldots, \hat{x}_n) P(z_1)$$
$$= \sum_{z_2} \sum_{z_1} P(y|z_1, z_2, x_1, \hat{x}_2, \ldots, \hat{x}_n) P(z_1)$$
$$P(z_2|z_1, x_1, \hat{x}_2, \ldots, \hat{x}_n)$$
$$= \sum_{z_2} \sum_{z_1} P(y|z_1, z_2, x_1, x_2, \hat{x}_3, \ldots, \hat{x}_n)$$
$$P(z_1) P(z_2|z_1, x_1)$$
$$- - - - $$
$$= \sum_{z_n} - - - \sum_{z_2} \sum_{z_1} P(y|z_1, \ldots, z_n, x_1, \ldots, x_n)$$
$$P(z_2|z_1, x_1) \ldots P(z_n|z_1, x_1, z_2, x_2, \ldots, z_{n-1}, x_{n-1})$$
$$= \sum_{z_1, \ldots, z_n} P(y|z_1, \ldots, z_n, x_1, \ldots, x_n)$$
$$\prod_{k=1}^{n} P(z_k|z_1, \ldots, z_{k-1}, x_1, \ldots, x_{k-1})$$

□

**Definition 1** *Any sequence $Z_1, \ldots, Z_n$ of covariates satisfying conditions (1) and (2) will be called "admissible" and any expression $P(y|\hat{x}_1, \hat{x}_2, \ldots, \hat{x}_n)$ which is identifiable by the criterion of Theorem 1 will be called G-identifiable.*

An immediate corollary of the definition above is

**Corollary 1** *A control problem is G-identifiable if it has an admissible sequence.*

G-identifiability is sufficient but not necessary for plan identifiability as defined above (see also Definition 3, Appendix I). The reasons are two fold. First, the completeness of the three inference rules used in the reduction of (3) is still a pending conjecture. Second, the $k^{th}$ step in the reduction of (3) refrains from conditioning on variables $Z_k$ that are descendants of $X_k$; namely, variables that may be affected by the action $do(X_k = x_k)$. In certain causal structures, the identifiability of causal effects requires that we condition on such variables [Pearl 1994].

Theorem 1 provides a declarative condition for plan identifiability. It can be used to ratify a proposed causal effect formula for a given plan, but does not provide an effective procedure for deriving such formulas, because the choice of each $Z_k$ is not spelled out procedurally; the possibility exists that some choices of $Z_k$, satisfying (1) and (2), might prevent us from continuing the reduction process even in cases where such reduction exists.

This is illustrated in Figure 3. Here $W$ is an admissible choice for $Z_1$, but if we make this choice we will not be able to complete the reduction, since no set $Z_2$ can be found that satisfies condition (2): $(Y \perp\!\!\!\perp X_2 | X_1, W, Z_2)_{G_{\underline{X_2}, \overline{X_1}}}$. In this case it would be wiser to choose $Z_1 = Z_2 = \emptyset$, which satisfies both: $(Y \perp\!\!\!\perp X_1 | \emptyset)_{G_{\underline{X_1}}}$, and $(Y \perp\!\!\!\perp X_2 | X_1, \emptyset)_{G_{\underline{X_2} \overline{X_1}}}$.

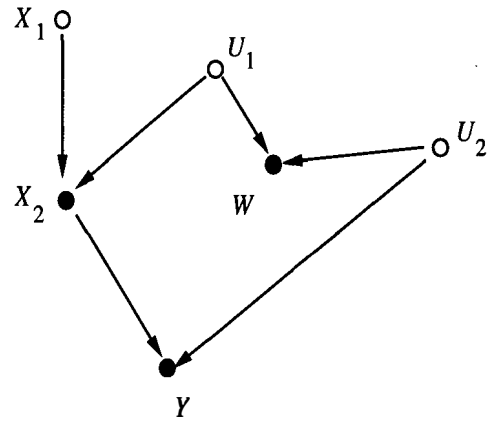

*Figure 3*: Illustrating an admissible choice $Z_1 = W$ that rules out any admissible choice for $Z_2$.

## 4    EVALUATION BY G-FORMULA

Let $L_k$ consist of *all* non descendants of $X_k$ which are descendants of $X_{k-1}$, including both observed and unobserved variables but exclusive of the controlled variables. Robins [1987] has shown, using counterfactual analysis, that

$$P(y|\hat{x}_1, \ldots, \hat{x}_n) =$$
$$\sum_{l_1, \ldots, l_n} P(y|l_1, \ldots, l_n, x_1, \ldots, x_n)$$
$$\prod_{k=1}^{n} P(l_k|l_1, l_2, \ldots, l_{k-1}, x_1, \ldots, x_{k-1}) \quad (5)$$

and named (5) the *G-formula* based on $L_1, \ldots, L_n$. One way of verifying (5) is to write the post-intervention distribution on all uncontrolled variables (using (18))

$$P(v_1, \ldots, v_n|\hat{x}_1, \ldots, \hat{x}_n) = \prod_i P(v_i|\mathbf{par}_{V_i})$$

$$= P(y|\mathbf{par}_Y) \prod_{k=1}^{n} P(l_k|\mathbf{par}_{L_k}) \qquad (6)$$

then take the marginal distribution on $Y$ by summing on the $l_k$'s. The identity of (6) and (5) follows from the independence

$$P(l_k|\mathbf{par}_{L_k}) = P(l_k|l_1,\ldots,l_{k-1},x_1,\ldots,x_{k-1}) \qquad (7)$$

Upon explicating the $l_k$'s in (5), we may find that some factors contain latent variables. When this happens we may try to use the conditional independencies encoded in the graph to eliminate those latent variables and, if we succeed, the plan would be identifiable and the resulting formula would give the desired causal effect.

Let us demonstrate this method on the example of Figure 1. The $L_k$ sequence is given by $L_1 = \{U_1\}$ and $L_2 = \{Z, U_2\}$. Substituting in the $G$-formula yields.

$$P(y|\hat{x}_1, \hat{x}_2) = \sum_{u_1,u_2,z} P(y|u_1, u_2, z, x_1, x_2) P(u_1)P(z, u_2|u_1, x_1)$$

Using the graph independencies $(Y \parallel \{Z, U_1\}|\{X_1, X_2, U_2\})_G$ and $(U_2 \parallel U_1|X_1)_G$, we get

$$P(y|\hat{x}_1, \hat{x}_2)$$
$$= \sum_{u_1,u_2,z} P(y|x_1, x_2, u_2)P(z|u_1, u_2, x)P(u_2|x_1)P(u_1)$$
$$= \sum_{u_2} P(y|x_1, x_2, u_2)P(u_2|x_1)$$
$$= \sum_{u_2} P(y|x_1, x_2, u_2) \sum_z P(u_2|x_1, z)P(z|x_1)$$
$$= \sum_{u_2} \sum_z P(y|x_1, x_2, z, u_2)P(u_2|x_1, z)P(z|x_1)$$
$$= \sum_z P(y|x_1, x_2, z)P(z|x_1)$$

which agrees with (4) under the admissible sequence $Z_1 = \emptyset\ Z_2 = Z$. Thus, by succeeding to eliminate the $U$ variable from the $G$-formula, we obtain a confirmation of plan identifiability together with the correct causal effect estimands.

The elimination method above still requires some search and algebraic skill. In addition, when the number of latent variables increases, the expressions tend to become rather involved. We now return to the problem of finding an admissible sequence, if one exists, thus eliminating the search altogether.

## 5 FINDING AN ADMISSIBLE SEQUENCE

The obvious way to avoid bad choices of covariates, like the one illustrated in Figure 3, is to insist on always choosing a "minimal" $Z_k$, namely, a set of covariates satisfying (2) having no proper subset which satisfies (2). However, since there are usually a large number of such minimal sets (see Figure 4), the question remains whether every choice of a minimal $Z_k$ is "safe", namely whether we can be sure that no choice of a minimal subsequence $Z_1,\ldots, Z_k$ will ever prevent us from finding an admissible $Z_{k+1}$, in case some admissible sequence $Z_1^*,\ldots, Z_n^*$ exists.

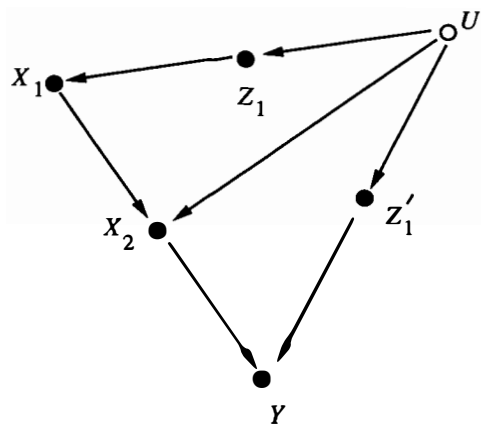

*Figure 4*: Illustrating non-uniqueness of minimal admissible sets: $Z_1$ and $Z_1'$ are each minimal and admissible.

The next result guarantees the "safety" of every minimal subsequence $Z_1,\ldots, Z_k$ and, hence, provides an effective test for $G$-identifiability.

**Theorem 2** *If there exists an admissible sequence $Z_1^*,\ldots, Z_n^*$, then for every minimally admissible subsequence $Z_1,\ldots, Z_{k-1}$ of covariates, there is an admissible set $Z_k$.*

**Proof:** The proof will be based on Lemmas 1 and 2 which are proved separately in Appendix II.

**Lemma 1** *For any DAG $G$ and any two disjoint subsets of nodes $X$ and $Y$, let the* ancestor-set *of $(X,Y)$, denoted $A(X,Y)$, be the set of nodes which have a descendant in either $X$ or $Y$.*

*The following two separation conditions hold for any sets of nodes $W$ and $Z$:*

$(Y \parallel X|Z, W \cap A(X,Y))_G$ whenever $(Y \parallel X|Z)_G$ \hfill (8)

$(Y \parallel X|W \cap A(X,Y))_G$ whenever $(Y \parallel X|W)_G$ \hfill (9)

Eq. (8) asserts that conditioning on nodes from an ancestral set can only create, never destroy independencies. Eq. (9) asserts that conditioning on all the nodes outside the ancestral set can only destroy, never create independencies.

**Lemma 2** *Denote by $G_k$ the subgraph $G_{\underline{X}_k, \overline{X}_{k+1}, \ldots, \overline{X}_n}$ of $G$, and let $A_k$ be the ancestral set of $(X_k, Y)$ in $G_k$. For any $j > 0$, $A_k$ is a subset of the ancestral set of $(X_k, Y)$ in $G_{k+j}$.*

We now prove Theorem 2 by contradiction. Suppose that $Z_1, \ldots, Z_{k-1}$ is minimally admissible sequence, and that no admissible set $Z_k$ exists. This means, in particular, that the set $Z_k = A_k \cap N_k$ is inadmissible, i.e.,

$$(Y \not\!\!\perp\!\!\!\perp X_k | X_1, \ldots, X_{k-1}, Z_1, \ldots, Z_{k-1}, A_k \cap N_k)_{G_k} \quad (10)$$

Now observe that no node in the sequence $Z_1, \ldots, Z_{k-1}$ can reside outside $A_k \cap N_k$. This is so because admissibility dictates $Z_i \in N_i$ and

$$(Y \perp\!\!\!\perp X_i | X_1, \ldots, X_{i-1}, Z_1, \ldots, Z_{i-1}, Z_i)_{G_i} \text{ for all } i < k \quad (11)$$

so, the lowest $i$ for which $Z_i$ contains a nonmember of $A_k$ will violate minimality (by (9)). Indeed, Lemma 2 insures that the violating $Z_i$ must also contain nonmembers of $A_i$ (in $G_i$), and (9) implies that if we remove all non-$A_i$ from a conditioning set, we do not destroy any separation. Moreover, since such a removal from $\{X_1, \ldots, X_{i-1}, Z_1, \ldots, Z_i\}$ will only affect $Z_i$, we can substitute $A_i$ for $Z_i$ in Eq. (11). This implies that $A_i$ satisfies (2) and $Z_i$ is non-minimal, which is a contradiction. We are now assured that $Z_1, \ldots, Z_{k-1}$ are in $A_k \cap N_k$. Likewise, since $\{X_1, \ldots, X_{k-1}\}$ is also in $A_k \cap N_k$, (10) can be rewritten as

$$(Y \not\!\!\perp\!\!\!\perp X_k | A_k \cap N_k) \quad (12)$$

To prove that (10) is false, contrast (12) with the assumption that there exists an admissible sequence $Z_1^*, \ldots, Z_n^*$. Let $Z^* = Z_k^* \bigcup_{i=1}^{k-1}(Z_i^* \cup \{X_i\})$. Admissibility states that (2) is satisfied by $Z_k = Z_k^*$, hence, $(Y \perp\!\!\!\perp X_k | Z^*)_{G_k}$. By (9), we can intersect the conditioning set $Z^*$ with $A_k$, yielding $(Y \perp\!\!\!\perp X_k | Z^* \cap A_k)$. Finally, since $Z^* \subseteq N_k$, we have

$$(Y \perp\!\!\!\perp X_k | Z^* \cap A_k \cap N_k)_{G_k} \quad (13)$$

But (12) and (13) together contradicts (8), because (8) asserts that whenever we add to the conditioning set members of $A_k$, we preserve independencies. QED

Theorem 2 now provides an effective decision procedure for testing $G$-identifiability:

**Corollary 2** *A control problem is $G$-identifiable if and only if the following algorithm exits with success:*

1. *Set $k = 1$*

2. *Choose any minimal $Z_k \in N_k$ satisfying (2),*

3. *If no such $Z_k$ exists, exit with failure. Else set $k = k + 1$,*

4. *If $k = n + 1$, exit with success, else go to step 2.*

From the proof of Theorem 2, it is obvious that we need not insist on choosing minimal $Z_k$. That requirement only insures that we do not step outside $A_k$ and spoil the selection of future subsets. In fact, Lemma 1 guarantees that if an admissible sequence exists, then $W_1, W_2, \ldots, W_n$ is such a sequence, where $W_k = A_k \cap N_k$. Accordingly, we can now rewrite Theorem 1 in terms of an explicit sequence of covariates.

**Theorem 3** *$P(y|\hat{x}_1, \ldots, \hat{x}_n)$ is $G$-identifiable if and only if the following condition holds for every $1 \le k \le n$*

$$(Y \perp\!\!\!\perp X_k | X_1, \ldots, X_{k-1}, W_1, W_2, \ldots, W_k)_{G_{\underline{X}_k, \overline{X}_{k+1}, \ldots, \overline{X}_n}}$$

*where $W_k = A_k \cap N_k$, namely, $W_k$ is the set of all covariates in $G$ that are both non-descendants of $X_k$ and have either $Y$ or $X_k$ as descendant. Moreover, when the condition above is satisfied the plan evaluates to*

$$P(y|\hat{x}_1, \ldots, \hat{x}_n) = \sum_{w_1, \ldots, w_n} P(y|w_1, \ldots, w_n, x_1, \ldots, x_n) \prod_{k=1}^{n} P(w_k|w_1, \ldots, w_{k-1}, x_1, \ldots, x_{k-1}) \quad (14)$$

## 6 GENERALIZATIONS

### 6.1 $Y$ AND $Z$ NON-DISJOINT

In practice, we will often be interested in a vector outcome $Y$ with components of $Y$ being ancestors of control variables $X_k$ for some $k$. For instance, in our AIDS example, we may be interested in survival $Y$ not only at a time after subjects have received treatment $X_2$ but also at a time after receiving treatment $X_1$ but before receiving $X_2$. If a component of $Y$ is both an ancestor of a control variable $X_k$ and of a later component of $Y$, it is necessary to regard the former component as a confounding variable that must be adjusted for to estimate the effect of the plan on $Y$. To do so, we no longer impose the assumption that $Y$ is a descendent of $X_n$ and that $Y$ and $Z$ are disjoint. Rather, we shall only require that $Y \subset Z$ where, henceforth, $Z$ represents all observed non-control variables. With this redefinition of $Y$, with the understanding that $(Y \perp\!\!\!\perp X|X)_G \ \forall X$, we prove below that

**Theorem 4** *Given $Y \subset Z$, Theorem 1 remains true.*

Further, under the above redefinitions of $Y$ and $Z$, we also obtain a natural generalization of Theorem 3. Let $Y_k^*$ be the subset of $Y$ that is not in $N_k$ and let $Y_k$ be the subset of $Y$ that is in $N_k$. Redefine $A_k$ to be the ancestral set of $(X_k, Y_k^*)$ in graph $G_k$. Robins and Pearl (1995) prove

**Theorem 5** : *Given $Y \subset Z$, Theorem 3 remains true with $W_k$ redefined to be $(A_k \bigcap N_k) \bigcup Y_k$.*

A key step in the proof of Theorem 5 is the following Lemma proved in [Robins & Pearl 1995].

**Lemma 3** *If a sequence $Z_k$ is G-admissible, then the sequence $Z_k \cup Y_k$ is also G-admissible.*

We shall also need this Lemma 3 in Section 6.3 below.

**Proof of Theorem 4:** Given a plan $x = (x_1, \ldots, x_n)$, define

$$h(y \mid x, \ell_1, \ldots, \ell_k) \equiv$$
$$\sum_{\ell_{k+1}, \ldots, \ell_n} P(y \mid \ell_1, \ldots, \ell_n, x_1, \ldots, x_n)$$
$$\prod_{m=k+1}^{n} P(\ell_m \mid \ell_1, \ldots, \ell_{m-1}, x_1, \ldots, x_{m-1}).$$

To prove Theorem 4, we shall use the following Lemma which is an easy consequence of the corollary to Theorem (AD.1) in Robins (1987).

**Lemma 4** *If, for each $k$, $Z_k \subset N_k$ and the expression*

$$\sum_{\ell_1, \ldots, \ell_k} h(y \mid x, \ell_1, \ldots, \ell_k)$$
$$P(\ell_1, \ldots, \ell_k \mid Z_1, \ldots, Z_k, X_k = x_k^*,$$
$$X_1 = x_1, \ldots, X_{k-1} = x_{k-1})$$

*does not depend on $x_k^*$ then Eq. (3) is true.*

One can also prove Lemma (4) directly by using induction on $n$ to show that the right hand side of Eq. (5) plus the premise of the lemma imply the right hand side of Eq. (3).

To complete the proof of Theorem 4, we shall show that (i) the premise of Lemma (4) is equivalent to the statement that

$$Y \amalg X_k \mid X_1 = x_1, \ldots, X_{k-1} = x_{k-1}, Z_1, \ldots, Z_k$$

when probabilities are computed under a particular joint distribution $P_{kx}$ for the variables $V$ in $G$ and (ii) $P_{kx}$ is represented by the DAG $G_k \equiv G_{\underline{X}_k, \overline{X}_{k+1}, \ldots, \overline{X}_n}$ [i.e., by definition, $P_{kx}(v) = \prod_j P_{kx}(v_j \mid pa_{jk})$ where $Pa_{jk}$ are the parents of $V_j$ on $G_k$ and $pa_{jk}$ is the value of $Pa_{jk}$ when $V = v$]. It then follows that Eqs. (1) and (2) imply the premise of Lemma (4), proving Theorem 4.

Let $P$ denote the distribution of variables $V$ on $G$. Now, given a plan $\hat{x} = (\hat{x}_1 \ldots \hat{x}_n)$, define $P_{kx}(v) = \prod_j P_{kx}(v_j \mid pa_{jk})$ where (i) if $V_j = X_m$ for some $m$, $m = k+1, \ldots, n$, then $P_{kx}(v_j \mid pa_{jk}) = 1$ if $v_j = x_m$, and (ii) if $V_j \neq X_m$ for $m = k+1, \ldots, n$, $P_{kx}(v_j \mid pa_{jk}) = P(v_j \mid pa_{jk})$ when $X_k$ is not a parent of $V_j$ on $G$, and $P_{kx}(v_j \mid pa_{jk}) = P(v_j \mid X_k = x_k, pa_{jk})$ when $X_k$ is a parent of $V_j$ on $G$. By construction, $P_{kx}$ is represented by the DAG $G_k$ and, therefore, $X_k \amalg Y \mid L_1, \ldots, L_k, X_1, \ldots, X_{k-1}$ under the distribution $P_{kx}$. Further, it is straightforward to calculate that (i) for any $x_k^*$, $h(y \mid x, \ell_1, \ldots, \ell_k) = P_{kx}(y \mid X_1 = x_1, \ldots, X_{k-1} = x_{k-1}, X_k = x_k^*, L_1 = \ell_1, \ldots, L_k = \ell_k)$, and (ii) the conditional distributions of $L_1, \ldots, L_k$ given $(Z_1, \ldots, Z_k, X_1, \ldots, X_k)$ are the same under $P$ and $P_{kx}$. Hence, the premise of Lemma (4) is equivalent to the conditional independence under the distribution $P_{kx}$ of $Y$ and $X_k$ given $(Z_1, \ldots, Z_k, X_1 = x_1, \ldots, X_{k-1} = x_{k-1})$.

We note that the premise of Lemma (4) is a non-graphical condition that is weaker than the graphical premise of Theorem 4 and yet implies identifiability by the G-formula based on $Z_1, \ldots, Z_n$. However, as a non-graphical condition, the premise of Lemma (4) is much more difficult to check that the graphical premise of Theorem 4.

## 6.2  $X_{k+j}$ NEED NOT BE A DESCENDENT OF $X_k$

In this subsection, we relax the assumption that $X_{k+j}$ is a descendent of $X_k$ for all $k, j > 0$. As in Sec. 6.1, $Z$ remains the set of all observed non-control variables. Given $X \subset V$, we say $X = (X_1, \ldots, X_n)$ is consistent ordering of $X$ in $G$ if, for each $k$, $X_k$ is a non-descendent of $\{X_{k+1}, \ldots, X_n\}$. Henceforth, given a consistent ordering of $X$, we redefine $N_k$ to be the set of observed non-control variables that are non-descendents of any element in the set $\{X_k, X_{k+1}, \ldots, X_n\}$. Robins and Pearl (1995) proved

**Theorem 6** *Given a consistent ordering $(X_1, \ldots, X_n)$ of $X$ with $X_k$ not necessarily an ancestor of $X_{k+j}$, Theorems 4 and 5 remain true.*

Theorem 6 is an immediate corollary of Theorems 4 and 5, and the following Theorem proved in Robins and Pearl (1995) characterizing arrows that can be added into and out of the $X_k$ without destroying Eqs. (1) and (2). Given a graph $G$, a consistent ordering $(X_1, \ldots, X_n)$ of $X$, and sets $Z_1, \ldots, Z_n, Z_k \subset N_k$, let graph $G^*$ be the graph in which, for each $k$, all arrows are included (i) from $X_k$ both to each member of the set $\{X_{k+1}, \ldots, X_n\}$ and to each variable (observed or unobserved) that is a descendent of some member of $\{X_{k+1}, \ldots, X_n\}$ and (ii) from each member of the set $Z_1 \bigcup \ldots \bigcup Z_k$ to $X_k$.

**Theorem 7** *Eqs. (1)-(2) hold for graph $G$ if and only if Eqs. (1)-(2) hold for graph $G^*$.*

Robins and Pearl (1995) show that the choice of consistent ordering for $X$ does matter. Specifically, they provide an example with $X = (X_a, X_b)$ bivariate in which both the ordering $(X_1, X_2) = (X_a, X_b)$ and the ordering $(X_1, X_2) = (X_b, X_a)$ are consistent orderings of $X$. However, $P(y \mid \hat{x})$ is only G-identifiable based on the ordering $X = (X_b, X_a)$.

### 6.3 VARIABLES THAT CAN BE DISCARDED

Eqs. (1)-(2) provide sufficient conditions for $G$-identification solely in terms of associations between observed variables. In the epidemiologic literature, sufficient conditions for $G$-identification are often expressed in terms of associations between unobserved and observed variables. For example, for the effect of a singleton action $X$ on $Y$, it is a standard result that an unobserved non-descendent of $X$, say $U$, is a "non-confounder given data on a non-descendent $Z_1$ of $X$" [i.e., $P(y \mid \hat{x})$ is $G$-identifiable based on $Z_1$] if either $U$ and $X$ are conditionally independent given $Z_1$ or if $U$ and $Y$ are conditionally independent given $(Z_1, X)$ [Miettinen & Cook 1981, Robins & Morgenstern 1987, Greenland & Robins 1986]. Extensions to compound actions are discussed in Robins (1986, Sec. 8 and Appendix F; 1989) and Robins et al. (1992, Sec. A2.13). The following theorem recasts Theorem 4 into this more familiar epidemiologic form. Given $Z_1, \ldots, Z_n$ with $Z_k \subset N_k$, let $U_k^*$ be all non-descendents of $\{X_k, \ldots, X_n\}$ (observed and unobserved) that are both non-control variables and are disjoint from $Z_1, \ldots, Z_k$. Robins and Pearl (1995) prove

**Theorem 8** *Suppose that, for each $k$, $Y_k \subset Z_k$. Then Eqs. (1)-(2) hold if and only if, for each $k$, $U_k^* = (U_{ak}^*, U_{bk}^*)$ for (possibly empty) disjoint sets $U_{ak}^*, U_{bk}^*$ satisfying*

(i)   $(U_{bk}^* \amalg X_k \mid Z_1, \ldots, Z_k, X_1, \ldots, X_{k-1})_{G_{\overline{X}_{k+1}, \ldots, \overline{X}_n}}$

and

(ii) $(U_{ak}^* \amalg Y_k^* \mid X_1, \ldots, X_k, Z_1, \ldots, Z_k, U_{bk}^*)_{G_{\overline{X}_{k+1}, \ldots, \overline{X}_n}}$.

Note that, in view of Lemma 3, the assumption $Y_k \subset Z_k$ is completely non-restrictive since we can always replace $Z_k$ by $Z_k \bigcup Y_k$ without destroying Eq. (1) or Eq. (2).

An important issue not treated in this paper is to derive sufficient conditions for the identification of $P(y \mid \hat{x})$ when $P(y \mid \hat{x})$ is not $G$-identifiable. Robins and Pearl (1995) provides sufficient conditions for identification of non-$G$-identifiable effects $P(y \mid \hat{x})$. When these criteria are satisfied, they provide a closed-form expression, called the composite-$G$-formula, for $P(y \mid \hat{x})$.

### Acknowledgment

Professor Pearl's research was partially supported by Air Force grant #AFOSR/F496209410173, NSF grant #IRI-9420306, and Rockwell/Northrop Micro grant #94-100 and Professor Robins' research was supported by NIH grant # RO1-AI32475.

# APPENDIX I

This appendix summarizes the basic definitions, notations and inference rules used in the body of the paper. Details and proofs can be found in [Pearl 1994, Pearl 1995].

Let $V = \{V_1, V_2, \ldots, V_n\}$ be the set of all variable in a directed acyclic graph (dag) $G$.

**Definition 2** (causal effect) Given two disjoint sets of variables, $X$ and $Y$, the causal effect of $X$ on $Y$, denoted $P(y|\hat{x})$, is a function from $X$ to the space of probability distributions on $Y$. For each realization $x$ of $X$, $P(y|\hat{x})$ gives the probability of $Y = y$ induced by the action $do(X = x)$.

If causal knowledge is organized as a set $T$ of structural equations

$$v_i = f_i(\mathbf{pa}_i, \epsilon_i) \quad i = 1, \ldots, n \quad (15)$$

where $\mathbf{pa}_i$ are the parents of $X_i$ in $G$, $f_i$ are (unspecified) deterministic functions and $\epsilon_i$ are mutually independent disturbances [Pearl & Verma 1991], then the joint distribution of the observed variables has the product from

$$P_T(v_1, \ldots, v_n) = \prod_i P(v_i|\mathbf{pa}_i) \quad (16)$$

independent of the $f_i$'s in $T$. In such a process-based theory, the effect of the action $do(V_j = v'_j)$ amounts to overruling the process governed by $f_i$ and substituting the process $V_j = v'_i$ instead. Consequently, the induced distribution in the mutilated theory $T_{v'_j}$ would be

$$P_{T_{v'_j}}(v_1, \ldots, v_n) \triangleq P_T(v_1, \ldots, v_2|\hat{v}'_j) \quad (17)$$

$$= \begin{cases} \prod_{i \neq j} P(v_i|\mathbf{pa}_i) = \frac{P(v_1, \ldots, v_n)}{P(v_j|\mathbf{pa}_j)} & \text{if } v_j = v'_j \\ 0 & \text{if } v_j \neq v'_j \end{cases}$$

independent of $T$. The partial product reflects the removal the factor $P(v_j|\mathbf{pa}_i)$ from the product of (16). Multiple actions result in the removal of the corresponding factors from (16).

**Definition 3** (identifiability) The causal effect of $X$ on $Y$ is said to be identifiable if the quantity $P(y|\hat{x})$ can be computed uniquely from any positive distribution of the observed variables. In other words $P_{T_1}(y|\hat{x}) = P_{T_2}(y|\hat{x})$ whenever $P_{T_1}(\mathbf{v}) = P_{T_2}(\mathbf{v}) > 0$.

Identifiability means that $P(y|\hat{x})$ can be estimated consistently from an arbitrarily large sample randomly drawn from the distribution of the observed variables.

The following theorem states the three basic inference rules used in the text.

**Theorem 9** Let $G$ be the directed acyclic graph associated with a causal model, and let $P(\cdot)$ stand for the probability distribution induced by that model. For any disjoint subsets of variables $X, Y, Z$, and $W$ we have:

**Rule 1** Insertion/deletion of observations

$$P(y|\hat{x}, z, w) = P(y|\hat{x}, w) \text{ if } (Y \perp\!\!\!\perp Z|X, W)_{G_{\overline{X}}} \quad (18)$$

**Rule 2** Action/observation exchange

$$P(y|\hat{x}, \hat{z}, w) = P(y|\hat{x}, z, w) \text{ if } (Y \perp\!\!\!\perp Z|X, W)_{G_{\overline{X}\underline{Z}}} \quad (19)$$

**Rule 3** Insertion/deletion of actions

$$P(y|\hat{x}, \hat{z}, w) = P(y|\hat{x}, w) \text{ if } (Y \perp\!\!\!\perp Z | X, W)_{G_{\overline{X}, \overline{Z(W)}}} \quad (20)$$

where $Z(W)$ is the set of $Z$-nodes that are not ancestors of any $W$-node in $G_{\overline{X}}$.

Each of the inference rules above follows from the basic interpretation of the "$\hat{x}$" operator as a replacement of the causal mechanism that connects $X$ to its pre-action parents by a new mechanism $X = x$ introduced by the intervening force. The result is a submodel characterized by the subgraph $G_{\overline{X}}$ (named "manipulated graph" in Spirtes et al. (1993)) which supports all three rules.

Rule 1 reaffirms $d$-separation as a valid test for conditional independence in the distribution resulting from the intervention $set(X = x)$, hence the graph $G_{\overline{X}}$. This rule follows from the fact that deleting equations from the system does not introduce any new dependencies among the remaining variables.

Rule 2 provides a condition for an external intervention $do(Z = z)$ to have the same effect on $Y$ as the passive observation $Z = z$. The condition amounts to $\{X \cup W\}$ blocking all back-door (i.e., spurious) paths from $Z$ to $Y$ (in $G_{\overline{X}}$), since $G_{\overline{X}\underline{Z}}$ retains all (and only) such paths.

Rule 3 provides conditions for introducing (or deleting) an external intervention $do(Z = z)$ without affecting the probability of $Y = y$. The validity of this rule stems, again, from simulating the intervention $do(Z = z)$ by pruning all links entering the variables in $Z$ (hence the graph $G_{\overline{XZ}}$).

**Corollary 3** A causal effect $q : P(y_1, \ldots, y_k | \hat{x}_1, \ldots, \hat{x}_m)$ is identifiable in a model characterized by a graph $G$ if there exists a finite sequence of transformations, each conforming to one of the inference rules in Theorem 9, which reduces $q$ into a standard (i.e., hat-free) probability expression involving observed quantities. □

## APPENDIX II

### Proof of Lemma 1

We will prove (8) by showing that if $(Y \perp\!\!\!\perp X | Z)_G$ holds, then augmenting $Z$ by any additional node $w \in A(X, Y)$ preserves the separation between $X$ and $Y$. Assume $w$ is an ancestor of $Y$. If $(Y \perp\!\!\!\perp X | Z)_G$ is true and $(Y \perp\!\!\!\perp X | Z, w)_G$ is false, then there must be a path between a node in $X$ and $Y$ that is blocked by $Z$ and become unblocked by $Z \cup \{w\}$. Let $\pi_1$ and $\pi_2$ be two parents of $w$ which became dependent by conditioning on $w$ and assume $\pi_1$ $d$-connects to $X$.

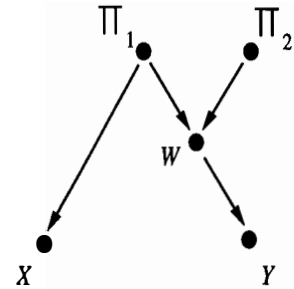

*Figure 5:*

Since all paths were blocked prior to conditioning on $w$, it must be that all paths from $w$ to $Y$ are blocked as well. But, since $w$ is an ancestor of $Y$, this means that some member of $Z$ resides on a directed path from $w$ to $Y$. This, however, means that $\pi_1$ and $\pi_2$ were not $d$-separated prior to conditioning on $w$; thus contradicting our basic assumption that conditioning on $w$ opened a new pathway between $X$ and $Y$. A symmetrical argument applies if $w$ is an ancestor of $X$ (or of both). Repeating the proof for each $w \in A(X, Y)$ completes the proof of (8).

To prove (9), we show that any path $p$ between $X$ and $Y$ that is blocked by $W$ will remain blocked when we remove from $W$ *all* nodes that are descendant of either $X$ or $Y$. Indeed, in order to unblock a path $p$ by removing nodes from $W$ some of the removed nodes must be non-colliders on $p$. Now, if $p$ is totally in $A(X, Y)$ no node on $p$ will be removed. On the other hand, if $p$ has some nodes outside $A(X, Y)$, it must have at least one collider $c$, such that $c$ and all its descendants are outside $A(X, Y)$. Therefore, when we remove from $W$ all non-ancestral nodes, we must leave $c$ and all its descendant unconditioned, hence $p$ must remain unblocked. QED

### Proof of Lemma 2

We shall first prove that any ancestor of $(Y, X_i)$ in $G_i$ is also an ancestor of $(Y, X_i)$ in $G_{i+j}$. If $t$ is an ancestor of $X_i$ in $G_i$ then clearly it must be an ancestor of $X_i$ in $G_{i+j}$; going from $G_{i+j}$ to $G_i$ does not affect any path incoming to $X_i$. Now assume that $t$ is an ancestor of $Y$ in $G_i$ but not in $G_{i+j}$. This can only happen if all paths (in $G$) from $t$ to $Y$ go through $X_{i+j}$ and get blocked in $G_i$ by removing the outgoing arrows from $X_{i+j}$. But any such path will be blocked in $G_i$ as well, because all incoming arrows to $X_{i+j}$ are removed in $G_i$, hence, $t$ cannot be an ancestor of $Y$ in $G_i$, which is a contradiction. We conclude that any ancestor of $Y$ in $G_i$ must also be an ancestor of $Y$ in $G_{i+1}$. Combining the two cases, completes the proof of Lemma 2.

(**Remark:** This proof relies on the assumption that each $X_{k+j}$ is an ancestor of $X_k$.)